\documentclass[12pt]{article}

\usepackage{lscape,multirow}
\usepackage{graphics, amsmath, amssymb, multirow, mathrsfs, graphicx, bm, colordvi, verbatim}
\usepackage{setspace, natbib}
\usepackage{caption, tabularx}
\usepackage{enumitem,colortbl}
\usepackage{url}
\usepackage{tablefootnote, ragged2e}
\usepackage{mathptmx}
\usepackage{pdflscape}
\usepackage{amsthm}
\usepackage{chngcntr}
\usepackage{changepage}
\usepackage{tikz}
\usepackage{tgpagella}
\usepackage{longtable}
\usepackage{pdflscape}
\usepackage{afterpage}
\usepackage{hyperref}
\usepackage{graphicx}
\usepackage[stable]{footmisc}
\usepackage{array} 
\newcolumntype{P}[1]{>{\centering\arraybackslash}p{#1}}
\hypersetup{
    colorlinks=true,       
    linkcolor=blue,          
    citecolor=blue,        
    filecolor=magenta,      
    urlcolor=blue           
}
\usepackage{changepage}
\usepackage{booktabs}

\usepackage{dcolumn}
\newcolumntype{d}[1]{D..{#1}} 

\usepackage[normalem]{ulem}

\newcolumntype{Y}{>{\centering\arraybackslash}X}
\newcolumntype{R}{>{\raggedleft\arraybackslash}X}

\usepackage{siunitx}
\usepackage[left=1in,right=1in,top=1in, bottom=1in]{geometry}

\usepackage{setspace}
\renewcommand{\thefootnote}{\fnsymbol{footnote}}

\begin{document}
\newgeometry{left=1in,right=1in,top=0in, bottom=1in} 
\begin{titlepage}
\title{\bf Instruction Tuning Chronologically Consistent Language Models
\thanks{\scriptsize Songrun He is at Washington University in St. Louis (\url{h.songrun@wustl.edu}). Linying Lv is at Washington University in St. Louis (\url{llyu@wustl.edu}). Asaf Manela is at Washington University in St. Louis (\url{amanela@wustl.edu}). Jimmy Wu is at Washington University in St. Louis (\url{jimmywu@wustl.edu}).}}
\author{Songrun He \quad\quad Linying Lv \quad\quad Asaf Manela \quad\quad Jimmy Wu}
        \date{First draft: July 2025. This draft: November 2025.}
        \maketitle

{\begin{abstract}\onehalfspacing
   We introduce a family of chronologically consistent, instruction-tuned large language models to eliminate lookahead bias. Each model is trained only on data available before a clearly defined knowledge-cutoff date, ensuring strict temporal separation from any post-cutoff data. The resulting framework offers (i) a simple, conversational chat interface, (ii) fully open, fixed model weights that guarantee replicability, and (iii) a conservative lower bound on forecast accuracy, isolating the share of predictability that survives once training leakage is removed. Together, these features provide researchers with an easy-to-use generative AI tool useful for a wide range of prediction tasks that is free of lookahead bias.
\end{abstract}}\small~

        \thispagestyle{empty}
\end{titlepage}
    \restoregeometry

\newpage
\setcounter{page}{1}
\renewcommand{\thefootnote}{\arabic{footnote}}
\setstretch{1.5}
\doublespacing

\newpage

\section{Introduction}\label{sec:intro}

Large language models (LLMs) have emerged as a transformative force in financial econometrics. Recent research extensively leverages LLM outputs for prediction and estimation tasks (see, for example: \citealp{lopez-lira_can_2023}; \citealp{chang2023ai}; \citealp{jha2024chatgpt}; \citealp{chen2025chatgpt}; \citealp{lv2025sell}). The novel capabilities and inherent intelligence of LLMs have enabled exploration of unstructured data and addressing previously unanswered questions.

However, as noted by \cite{sarkar2024lookahead} and \cite{ludwig2025large}, most prediction problems with generative AI face methodological challenges of lookahead bias. The reason is that LLMs are pretrained on a vast corpus of text data that incorporates future information relative to the prediction task. This leads to lookahead bias when the model's knowledge cutoff $\tau$ extends beyond the prediction time, as a training leakage term then emerges in the loss function.

Several papers have introduced robust methods for isolating temporal information. For example, \cite{glasserman2023assessing} and \cite{engelberg2025entity} develop a systematic entity masking approach so that LLMs cannot recognize firms. Alternatively, researchers have explored pretraining LLMs from scratch, meticulously curating the training data to control the models' inherent knowledge (e.g., \citealp{sarkar_storieslm_2024}; \citealp{he2025chronologically}).

While pretraining an LLM from scratch appears to be a more comprehensive solution for preventing the model from accessing future knowledge, as opposed to methods that attempt to make it forget learned information or infer from context, its implementation presents two primary challenges. First, ensuring chronological consistency requires excluding a substantial amount of future training text, which may compromise the model's performance. Second, integrating this approach, especially within a model embedding and machine learning pipeline, is both technically sophisticated and computationally expensive.

In this paper, we offer another solution more accessible to the social science research community: the first instruction-following chat model free of lookahead bias. Specifically, during both the pretraining and instruction finetuning stages, we carefully curate the dataset to prevent the model from seeing future knowledge. For example, $\text{ChronoGPT-Instruct}_{\tau}$ (where $\tau$ is in $\{1999,2000,2001,...,2024\}$ ) never accesses any knowledge that emerged or became economically salient after $\tau$. For any evaluation set post-$\tau$, the model has perfect temporal separation from the evaluation set.

Despite the significant effort in developing the ChronoGPT-Instruct series, certain challenges in its design and implementation are important to highlight. A fundamental tradeoff exists between maintaining robust chat capabilities and ensuring strict chronological consistency. To illustrate, a Qwen-1.5-1.8B-Chat model of similar parameter size is pretrained on 2.2 trillion tokens, approximately 31 times the 70 billion tokens seen by our base model. Nevertheless, even the earliest ChronoGPT-Instruct models achieve above 12\% win rates in the Alpaca instruction-following evaluation, demonstrating their practical utility despite data constraints. Additionally, while our prompt-based filtering algorithm is designed to be highly effective, we acknowledge it is not theoretically flawless. However, in a rigorous validation test we conduct, ChronoGPT-Instruct models consistently fail to predict future presidents or major events, showing no signs of training leakage.

The primary contribution of our model is to serve as a useful tool for conducting lookahead bias-free robustness tests in various prediction problems. We publicly release our ChronoGPT-instruct models and instruction-finetuning data to support the research community at: \url{https://huggingface.co/manelalab}. While ChronoGPT-Instruct does not offer a perfect solution that simultaneously eliminates training leakage and preserves state-of-the-art language abilities, it allows for establishing a conservative lower bound of predictive power, providing a clearer understanding of true model performance. In a prompt-based trading portfolio example we provide, if ChronoGPT-Instruct is considered a lookahead bias-free counterpart to larger models such as Qwen-1.5-1.8B-Chat and Llama-3.2-3B-Instruct (up to twice the parameter count and trained on far more data), our finding implies that at least 54\% of the observed news return predictability persists without leakage. The remaining discrepancy in Sharpe ratios (e.g., between 0.95 and 1.76) likely stems from a combination of differences in model capabilities and lookahead bias in the comparison model.

\section{Methodology and Data}\label{sec:method}

In this section, we describe our instruction finetuning methodology designed to enforce no training leakage, followed by details of our data curation process and the datasets used for both instruction finetuning and return prediction tasks.

\subsection{Instruction Finetuning}

Our instruction finetuning pipeline is designed to satisfy the \emph{no-training-leakage} contract of \citet{ludwig2025large}. The contract formalizes the idea that any text used for evaluating a model must be statistically independent of the text used to train it. We first restate the contract in a two-stage setting of pretraining and instruction finetuning (IFT), and then show how our data curation enforces each of its requirements.

\noindent\textbf{Two disjoint training corpora.} Let
\begin{equation}
    t^{\mathrm{pre}}(\tau)=\left\{\sigma \in \Sigma^*: \operatorname{date}(\sigma) \leq \tau\right\}, \quad t^{\mathrm{ift}}(\tau)=\left\{\sigma \in \Sigma_{\text {inst }}^*: \operatorname{date}(\sigma) \leq \tau\right\},
\end{equation}
where $\tau$ is the knowledge cutoff of our vintage models. For any text piece $r$ define stage-specific indicators
\begin{equation}
t_r^{\mathrm{pre}}=\mathbf{1}\big(r \in t^{\mathrm{pre}}(\tau)\big), \quad t_r^{\mathrm{ift}}=\mathbf{1}\big(r \in t^{\mathrm{ift}}(\tau)\big)
\end{equation}
and let
\begin{equation}\label{equa:combine}
t_r=\max \left\{t_r^{\mathrm{pre}}, t_r^{\mathrm{ift}}\right\} \in\{0,1\}
\end{equation}
denote membership in the combined training set.

\noindent\textbf{Loss decomposition with a leakage term.} Consider an evaluation sample $R_{>\tau}$ consisting exclusively of documents dated after $\tau$ and let $D_r=\mathbf{1}\left(r \in R_{>\tau}\right)$. With loss function $\ell(\cdot, \cdot)$ and model prediction $\hat{m}(r ; t)$, the expectation of $\hat{L}_\tau$ can be written as
\begin{equation}\label{equa:leakeage} E\left[\hat{L}_\tau\right]=\underbrace{E\left[\ell\left(Y_r, \hat{m}(r ; t)\right)\right]}_{\text {true out-of-sample loss }}-\underbrace{E\left[D_r\left(\frac{q_{T \mid D}\left(t_r\right)}{q_T\left(t_r\right)}-1\right) \ell\left(Y_r, \hat{m}(r ; t)\right)\right]}_{\text {leakage term }},
\end{equation}
where
\begin{equation}
q_T\left(t_r\right)=\operatorname{Pr}\left(t_r=1\right), \quad q_{T \mid D}\left(t_r\right)=\operatorname{Pr}\left(t_r=1 \mid D_r=1\right)
\end{equation}
are the \emph{unconditional} and \emph{conditional} probabilities that $r$ appears in the training set, respectively. The second expectation in (\ref{equa:leakeage}) is the leakage term; it vanishes if and only if
\begin{equation}
\forall r: \frac{q_{T \mid D}\left(t_r\right)}{q_T\left(t_r\right)}=1,
\end{equation}
which is the contract's independence condition.

\noindent\textbf{Stage-wise sufficiency.} Because the overall indicator $t_r$ in (\ref{equa:combine}) is the union of two disjoint events, independence is guaranteed once it holds separately for pretraining and IFT:
\begin{equation}\label{equa:no_leakage}
\forall r: \frac{q_{T \mid D}\left(t_r^{\mathrm{pre}}\right)}{q_T\left(t_r^{\mathrm{pre}}\right)}=1 \quad \text { and } \quad \frac{q_{T \mid D}\left(t_r^{\mathrm{ift}}\right)}{q_T\left(t_r^{\mathrm{ift}}\right)}=1.
\end{equation}

For the pretraining stage, we use the vintage ChronoGPT in \citet{he2025chronologically} as our base model. The corpus $T_\tau^{\text {pre }}$ is built from historical web snapshots, archived news, and scientific literature. Every document carries a verifiable publication timestamp, and any text dated after $\tau$ is discarded. Hence for each post-knowledge cutoff evaluation item $r$ we have $t_r^{\text {pre }}=0$, leading to
\begin{equation}
q_T\left(t_r^{\mathrm{pre}}\right)=0, \quad q_{T \mid D}\left(t_r^{\mathrm{pre}}\right)=0,
\end{equation}
so the first equality in (\ref{equa:no_leakage}) holds.

\vspace{0.1in}
\noindent
{\small
\begin{tabular}{|p{16cm}|}
\hline
\textbf{Prompt:} You are provided with a user-assistant interaction. Your task is to determine whether the conversation contains any information that would have been unavailable or irrelevant prior to the year 2000. \\
Specifically, indicate whether the message includes any direct or indirect reference to:\\
1. A concept, company, product, technology, event, online review, or terminology that was created, discovered, or publicly introduced after 1999, or\\
2. A subject that only gained significant economic, cultural, scientific, or technological relevance after 1999, even if it existed before that date.\\
If such a reference is present anywhere in the conversation, return: 1\\
If the conversation is entirely composed of content that could have been generated using only knowledge available prior to 2000, return: 0\\
Clarifications:\\
- For conversations evaluated as low quality, also assign a label of 1.\\
- In cases of uncertainty or ambiguity, adopt a conservative approach and assign a label of 1.\\
- References to post-1999 entities such as GPT models, Kubernetes, TikTok, blockchain, COVID-19, Tesla, or similar modern constructs are strong indicators of a label of 1.\\
Return your answer strictly as a JSON object with the following fields:\\
- "label": either 0 or 1\\
- "confidence": a number from 0 to 10 (higher means more certain)\\
- "suspected term": a brief phrase (1-3 words) that triggered your label decision, or "none" if label=0\\
Example output: \{"label": 1, "confidence": 9, "suspected term": "GPT-3"\}\\
Here is the message:\{conversation\}\\
\hline
\end{tabular}
}
\vspace{0.1in}

We conduct a prompt filtering algorithm for the instruction-finetuning stage. Candidate instruction-response pairs are screened with an LLM classifier. The classifier, implemented with ChatGPT-4.1, receives the following prompt and returns a binary label indicating whether the dialogue contains any knowledge that emerged or became economically salient after $\tau$. 

Only pairs receiving the label 0 are admitted to $t^{\text {ift}}(\tau)$. Consequently, for every evaluation item $r$ dated after the cut-off, $t_r^{\mathrm{ift}}=0$ and the second equality in (\ref{equa:no_leakage}) also holds. We conduct a validation test for the independence condition in section \ref{sec:validation}.

\subsection{Data}

This section introduces the public user-assistant interaction dataset we use for instruction finetuning, and the financial newswire data we use for return prediction.

\subsubsection{Instruction Finetuning Data}

The instruction-finetuning corpus comprises over 425,000 prompt–response pairs drawn from three public resources and arranged as a curriculum that grows in cognitive load and sequence length. We start with simple, short tasks from \citet{raschka2024build}, like spelling checks or basic math. Then, we add medium-length prompts from the \citet{wang2022self} dataset generated from the GPT-3 through the self-instruct technique. Finally, we include the broad AllenAI's Tulu-3 SFT mixture created by \citet{lambert2024tulu}.

All entries are filtered to (i) exclude non-English records and code snippets, and (ii) satisfy a temporal-knowledge screen: each example is classified by GPT-4.1 and only retained when the model assigns label 0 (“knowledge available pre-2000”) with the maximum confidence score of 10. Table \ref{tab:sftdata} summarizes the resulting dataset.

\begin{table}[!htb]
\begin{center}
\begin{tabularx}{\textwidth}{@{\hskip\tabcolsep\extracolsep\fill}lrrr}
\toprule
\textbf{Stage} & \textbf{SFT data} & \textbf{Number of examples} & \textbf{Average conversation length} \\
\midrule
1 & LLMs-from-scratch             & 1,097  & 102  \\
2 & GPT-3 self-generated          & 67,136 & 183  \\
3 & Tulu-3 SFT mixture        & 356,886 & 2,513 \\
\bottomrule
\end{tabularx}
\end{center}
\caption{Instruction Finetuning Datasets}
\label{tab:sftdata}
\end{table}

We then format these entries as inputs to ChronoGPT using Alpaca-style prompt formatting. Below is an example entry passed to the LLM:
\begin{verbatim}
{ 
  Below is an instruction that describes a task. Write a response that
  appropriately completes the request.
    ### Instruction:
    Identify the correct spelling of the following word.
    ### Input:
    Ocassion
    ### Response:
    The correct spelling is 'Occasion.'
}
\end{verbatim}

\subsubsection{Financial Newswire Data}

We use the Dow Jones Newswire dataset, a real-time newswire providing extensive coverage of financial markets from January 2007 to July 2023. This dataset includes news headlines, full article texts, and precise display timestamps. Following \citet{he2025chronologically}, we focus on firm-specific news, aggregating all relevant headlines for each firm within a trading day. Finally, we merge this news data with CRSP close-to-close returns on day t+1 to examine the predictability of stock returns.

\section{Results}\label{sec:empirical}

\subsection{Instruction Following Evaluation}

We instruction-finetune a series of models as \text{ChronoGPT-Instruct}-{1999}, \text{ChronoGPT-Instruct}-{2005}, \text{ChronoGPT-Instruct}-{2010}, \text{ChronoGPT-Instruct}-{2015}, \text{ChronoGPT-Instruct}-{2020}, and \text{ChronoGPT-Instruct}-{2024}, each starting from the corresponding ChronoGPT vintage model in \citet{he2025chronologically}. All vintages are finetuned with the standard masked cross-entropy for next-token prediction, formally defined as
\begin{equation}
\mathcal{L}=-\frac{1}{N} \sum_{t=1}^N \log p_\theta\left(y_t \mid \mathbf{x}_{<t}\right),
\end{equation}
where the model parameters $\theta$ are optimized to maximize the probability $p_\theta\left(y_t \mid \mathbf{x}_{<t}\right)$ of generating the true token $y_t$ at position $t$ within a sequence of length $N$. This objective rewards the model for accurately predicting the subsequent token in a response.

\begin{figure}[!htb]
\begin{footnotesize}
\begin{center}
\includegraphics[width=\linewidth]{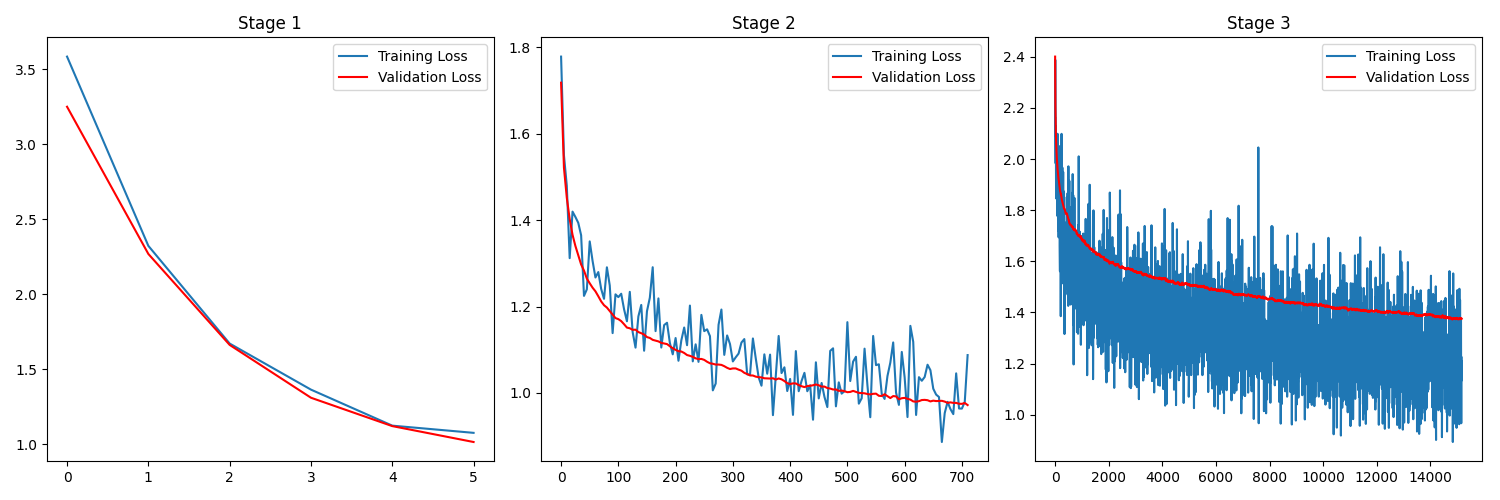}
\end{center}
\end{footnotesize}
\caption{Training Loss and Validation Loss of Instruction Finetuning}
\label{fig:instructuin_finetuning}
\bigskip
\small
The figure shows the training dynamics across three stages of supervised finetuning (SFT) on the ChronoGPT-1999 model. Stage 1 uses LLMs-from-scratch data. Stage 2 uses GPT-3 self-generated data. Stage 3 uses Tulu-3-SFT mixture.
\end{figure}

Every SFT stage logs the token-level cross-entropy on a 5\% hold-out split that is never seen by the optimizer. Figure~\ref{fig:instructuin_finetuning} plots these losses across three stages of training, showing the characteristic rapid decline in early steps followed by more gradual improvement. The steep initial drop reflects rapid adaptation to the instruction-following format, while later stages show continued learning from the curriculum order.

\begin{figure}[!htb]
\begin{footnotesize}
\begin{center}
\includegraphics[width=\linewidth]{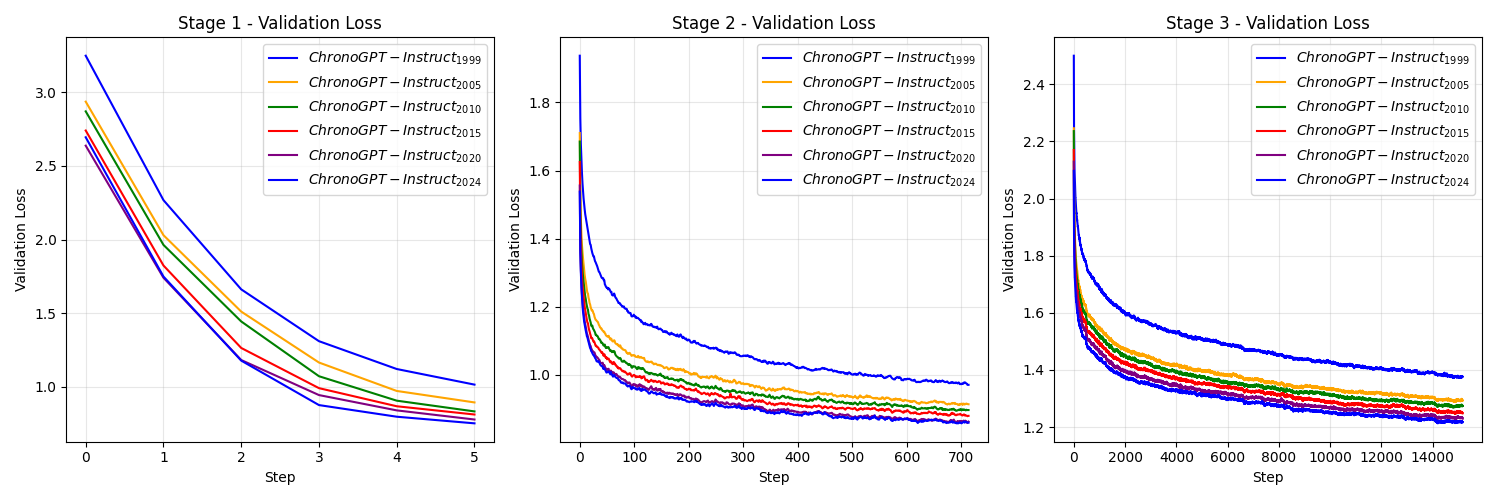}
\end{center}
\end{footnotesize}
\caption{Validation Loss of Instruction Model Vintages}\label{fig:val}
\bigskip
\small
The figure compares the validation loss across six vintage models (1999, 2005, 2010, 2015, 2020, 2025) for the three curriculum stages: Stage 1 (LLMs-from-scratch, simple tasks), Stage 2 (GPT-3 self-generated, medium complexity), and Stage 3 (Tulu-3 mixture, complex conversations).
\end{figure}

Figure~\ref{fig:val} compares the validation losses across all five vintages for each training stage. Consistent with the language-model results in \citet{he2025chronologically}, we observe systematic improvements from earlier to later vintages, with the 1999 model showing the highest validation loss and more recent vintages achieving lower losses across all three stages.

\begin{figure}[!htb]
\begin{footnotesize}
\begin{center}
\includegraphics[width=0.8\linewidth]{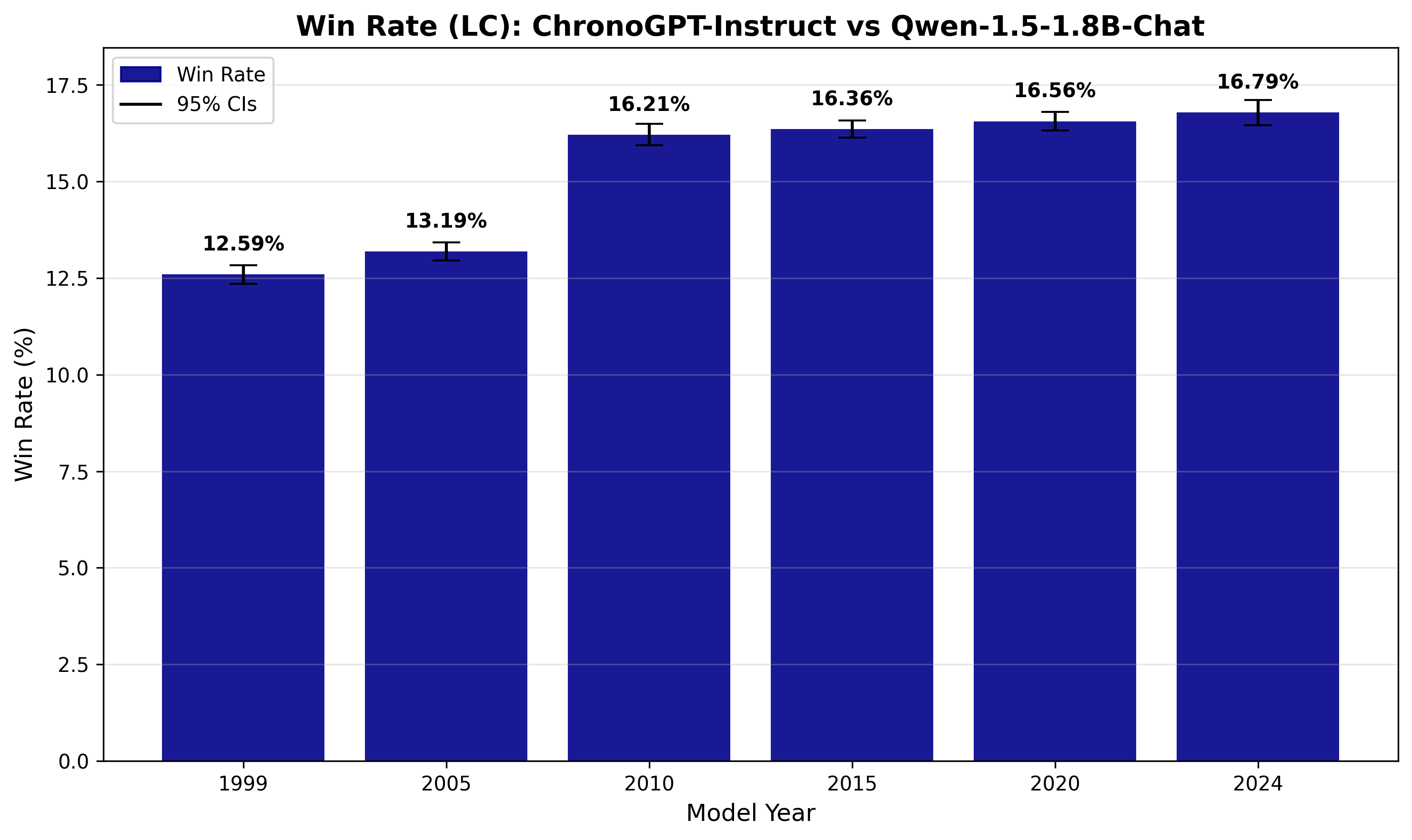}
\end{center}
\end{footnotesize}
\caption{Alpaca Evaluation for ChronoGPT-Instruct}\label{fig:alpaca}
\bigskip
\small
This figure shows head-to-head win rates using length-controlled evaluation across model vintages from 1999 to 2024. The evaluation set is from AlpacaFarm. The benchmark model is Qwen-1.5-1.8B-Chat.
\end{figure}

While validation loss is a useful proxy, it conflates syntax prediction with pragmatic instruction-following. We therefore evaluate using a head-to-head comparison against Qwen-1.5-1.8B-Chat on Alpaca length-controlled (LC) evaluation. For each instruction in the AlpacaEval dataset, we generated outputs from both our model and the reference model. These pairs of outputs were then presented to an automatic evaluator, which determined a preference. A 50\% win rate indicates that the evaluated model performs comparably to the reference model.

Figure~\ref{fig:alpaca} reports the resulting win rates across vintages. $\text{ChronoGPT-Instruct}_{1999}$ achieves a 12.59\% win rate. Performance increases to 13.19\% in 2005, 16.21\% in 2010, and peaks at 16.79\% in 2024. This steady improvement demonstrates that more recent training data consistently enhances instruction-following capabilities, with the 2024 vintage showing the strongest performance. Despite these gains, the overall low win rates largely stem from the significant disparity in pretraining data volume: the reference Qwen-1.5-1.8B-Chat model was pretrained on approximately 31 times more tokens than our base ChronoGPT models.

\subsection{Chronological Consistency Validation} \label{sec:validation}

\begin{table}[!htb]
    \resizebox{\textwidth}{!}{
    \begin{tabular}{lccccccccc}
        \toprule
        & \multicolumn{6}{c}{Election year} & & \multicolumn{2}{c}{Accuracy}\\
        \cmidrule{2-7} \cmidrule{9-10}
         & 1992 & 2000 & 2008 & 2016 & 2020 & 2024 & & Pre-cutoff & Post-cutoff \\
        \midrule
        Correct output & Bill Clinton & George W. (Bush) & Barack Obama & Donald Trump & Joe Biden & Donald Trump\\
        \midrule
        GPT-2              & \textcolor{blue}{\textbf{Bill Clinton}}  & Bill Clinton  & \textcolor{blue}{\textbf{Barack Obama}}  & \textcolor{blue}{\textbf{Donald Trump}}  &\cellcolor{lightgray}George W.  &\cellcolor{lightgray}George W. & & $3/4$ & \cellcolor{lightgray}$0/2$\\
        GPT-2 XL             & \textcolor{blue}{\textbf{Bill Clinton}}  & \textcolor{blue}{\textbf{George W.}}  & \textcolor{blue}{\textbf{Barack Obama}}  & \textcolor{blue}{\textbf{Donald Trump}}  &\cellcolor{lightgray}James A.  &\cellcolor{lightgray}James Mattis & & $4/4$ & \cellcolor{lightgray}$0/2$\\
        Llama-3.2-3B-Instruct & \textcolor{blue}{\textbf{Bill Clinton}}  & \textcolor{blue}{\textbf{George W.}}  & \textcolor{blue}{\textbf{Barack Obama}}  & \textcolor{blue}{\textbf{Donald Trump}}  & \textcolor{blue}{\textbf{Joe Biden}}  &\cellcolor{lightgray}R. & & $5/5$ & \cellcolor{lightgray}$0/1$\\
        Qwen-1.5-1.8B-Chat & \textcolor{blue}{\textbf{Bill Clinton}}  & \textcolor{blue}{\textbf{George W.}}  & \textcolor{blue}{\textbf{Barack Obama}}  & \textcolor{blue}{\textbf{Donald Trump}}  & \textcolor{blue}{\textbf{Joe Biden}}  &\cellcolor{lightgray}Kamala & & $5/5$ & \cellcolor{lightgray}$0/1$\\
        \midrule
        $\text{ChronoGPT-Instruct}_{\text{Realtime}}$ & --- & \cellcolor{lightgray} Bill Clinton & \cellcolor{lightgray} George W. & \cellcolor{lightgray} John F. & \cellcolor{lightgray} elect2019: & \cellcolor{lightgray} Joe Biden & & --- & \cellcolor{lightgray} 0/5 \\
        \midrule
        $\text{ChronoGPT-Instruct}_{1999}$   &  \textcolor{blue}{\textbf{Bill Clinton}}  & \cellcolor{lightgray}Bill Clinton  & \cellcolor{lightgray}Bill Clinton  & \cellcolor{lightgray}Clinton\textbackslash nT  & \cellcolor{lightgray}Obama\textbackslash nT  & \cellcolor{lightgray}John F. & &  $1/1$ & \cellcolor{lightgray} $0/5$\\
        $\text{ChronoGPT-Instruct}_{2000}$   & \textcolor{blue}{\textbf{Bill Clinton}}  & Bill Clinton  & \cellcolor{lightgray}Bill Clinton  & \cellcolor{lightgray}Bill Clinton  & \cellcolor{lightgray}Bill Clinton  & \cellcolor{lightgray}John W. &  & $1/2$ & \cellcolor{lightgray} $0/4$\\
        $\text{ChronoGPT-Instruct}_{2001}$   & \textcolor{blue}{\textbf{Bill Clinton}}  & \textcolor{blue}{\textbf{George W.}}  & \cellcolor{lightgray}Bill Clinton  & \cellcolor{lightgray}George W.  & \cellcolor{lightgray}George W.  & \cellcolor{lightgray}Putin\textbackslash nT & &  $2/2$ & \cellcolor{lightgray} $0/4$\\
        $\text{ChronoGPT-Instruct}_{2002}$  & \textcolor{blue}{\textbf{Bill Clinton}}  & \textcolor{blue}{\textbf{George W.}}  & \cellcolor{lightgray}Bill Clinton  & \cellcolor{lightgray}George W. & \cellcolor{lightgray}George W.  & \cellcolor{lightgray}George W. & &  $2/2$ & \cellcolor{lightgray} $0/4$\\
        $\text{ChronoGPT-Instruct}_{2003}$  & \textcolor{blue}{\textbf{Bill Clinton}}  & \textcolor{blue}{\textbf{George W.}}  & \cellcolor{lightgray}George W.  & \cellcolor{lightgray}George W. & \cellcolor{lightgray}George W.  & \cellcolor{lightgray}George W. & &  $2/2$ & \cellcolor{lightgray} $0/4$\\
        $\text{ChronoGPT-Instruct}_{2004}$  & \textcolor{blue}{\textbf{Bill Clinton}}  & \textcolor{blue}{\textbf{George W.}}  & \cellcolor{lightgray}George W.  & \cellcolor{lightgray}George W. & \cellcolor{lightgray}George W.  & \cellcolor{lightgray}Putin\textbackslash nT  & &  $2/2$ & \cellcolor{lightgray} $0/4$\\
        $\text{ChronoGPT-Instruct}_{2005}$  & \textcolor{blue}{\textbf{Bill Clinton}}  & \textcolor{blue}{\textbf{George W.}}  & \cellcolor{lightgray}George W.  & \cellcolor{lightgray}George W. & \cellcolor{lightgray}George W.  & \cellcolor{lightgray}George W. & &  $2/2$ & \cellcolor{lightgray} $0/4$\\
        $\text{ChronoGPT-Instruct}_{2006}$  & \textcolor{blue}{\textbf{Bill Clinton}}  & \textcolor{blue}{\textbf{George W.}}  & \cellcolor{lightgray}George W.  & \cellcolor{lightgray}George W. & \cellcolor{lightgray}George W.  & \cellcolor{lightgray}George W. & &  $2/2$ & \cellcolor{lightgray} $0/4$\\
        $\text{ChronoGPT-Instruct}_{2007}$  & \textcolor{blue}{\textbf{Bill Clinton}}  & \textcolor{blue}{\textbf{George W.}}  & \cellcolor{lightgray}George W.  & \cellcolor{lightgray}George W. & \cellcolor{lightgray}George W.  & \cellcolor{lightgray}George W. & &  $2/2$ & \cellcolor{lightgray} $0/4$\\
        $\text{ChronoGPT-Instruct}_{2008}$  & \textcolor{blue}{\textbf{Bill Clinton}}  & \textcolor{blue}{\textbf{George W.}}  & George W.  & \cellcolor{lightgray}George W. & \cellcolor{lightgray}George W.  & \cellcolor{lightgray}George W. & &  $2/3$ & \cellcolor{lightgray} $0/3$\\
        $\text{ChronoGPT-Instruct}_{2009}$  & \textcolor{blue}{\textbf{Bill Clinton}}  & \textcolor{blue}{\textbf{George W.}} & George W.  & \cellcolor{lightgray}George W. & \cellcolor{lightgray}George W.  & \cellcolor{lightgray}George W. & &  $2/3$ & \cellcolor{lightgray} $0/3$\\
        $\text{ChronoGPT-Instruct}_{2010}$  & \textcolor{blue}{\textbf{Bill Clinton}}  & \textcolor{blue}{\textbf{George W.}} & \textcolor{blue}{\textbf{Barack Obama}}  &\cellcolor{lightgray}Barack Obama  &\cellcolor{lightgray}Bill Gates  &\cellcolor{lightgray}Bill Gates &  &  $3/3$ & \cellcolor{lightgray} $0/3$\\
        $\text{ChronoGPT-Instruct}_{2011}$  & \textcolor{blue}{\textbf{Bill Clinton}}  & \textcolor{blue}{\textbf{George W.}} & \textcolor{blue}{\textbf{Barack H.}}  &\cellcolor{lightgray}John F.  &\cellcolor{lightgray}Bill Gates  &\cellcolor{lightgray}George W. &  &  $3/3$ & \cellcolor{lightgray} $0/3$\\
        $\text{ChronoGPT-Instruct}_{2012}$  & \textcolor{blue}{\textbf{Bill Clinton}}  & \textcolor{blue}{\textbf{George W.}} & \textcolor{blue}{\textbf{Barack Obama}}  &\cellcolor{lightgray}Barack Obama  &\cellcolor{lightgray}Bill Gates  &\cellcolor{lightgray}George W. &  &  $3/3$ & \cellcolor{lightgray} $0/3$\\
        $\text{ChronoGPT-Instruct}_{2013}$  & \textcolor{blue}{\textbf{Bill Clinton}}  & \textcolor{blue}{\textbf{George W.}} & \textcolor{blue}{\textbf{Barack Obama}}  &\cellcolor{lightgray}John F.  &\cellcolor{lightgray}Bill Gates  &\cellcolor{lightgray}George W. &  &  $3/3$ & \cellcolor{lightgray} $0/3$\\
        $\text{ChronoGPT-Instruct}_{2014}$  & \textcolor{blue}{\textbf{Bill Clinton}}  & \textcolor{blue}{\textbf{George W.}} & \textcolor{blue}{\textbf{Barack Obama}}  &\cellcolor{lightgray}John F.  &\cellcolor{lightgray}George W.  &\cellcolor{lightgray}George W. &  &  $3/3$ & \cellcolor{lightgray} $0/3$\\
        $\text{ChronoGPT-Instruct}_{2015}$  & \textcolor{blue}{\textbf{Bill Clinton}}  & \textcolor{blue}{\textbf{George W.}} & George W. &\cellcolor{lightgray} John F.  &\cellcolor{lightgray}George W.  &\cellcolor{lightgray}George W. &  &  $2/3$ & \cellcolor{lightgray} $0/3$\\
        $\text{ChronoGPT-Instruct}_{2016}$  & \textcolor{blue}{\textbf{Bill Clinton}}  & \textcolor{blue}{\textbf{George W.}} & George W.  & Martin Luther King &\cellcolor{lightgray}George W.  &\cellcolor{lightgray}George W. &  &  $2/4$ & \cellcolor{lightgray} $0/2$\\
        $\text{ChronoGPT-Instruct}_{2017}$  & \textcolor{blue}{\textbf{Bill Clinton}}  & \textcolor{blue}{\textbf{George W.}} & George W.  & \textcolor{blue}{\textbf{Donald Trump}}   &\cellcolor{lightgray}John F. & \cellcolor{lightgray}John Kasich &  &  $3/4$ & \cellcolor{lightgray} $0/2$\\
        $\text{ChronoGPT-Instruct}_{2018}$  & \textcolor{blue}{\textbf{Bill Clinton}}  & \textcolor{blue}{\textbf{George W.}} & George W.  & \textcolor{blue}{\textbf{Donald Trump}}   & \cellcolor{lightgray}Donald Trump & \cellcolor{lightgray}George W. &  &  $3/4$ & \cellcolor{lightgray} $0/2$\\
        $\text{ChronoGPT-Instruct}_{2019}$  & \textcolor{blue}{\textbf{Bill Clinton}}  & \textcolor{blue}{\textbf{George W.}} & George W.  & \textcolor{blue}{\textbf{Donald Trump}}   & \cellcolor{lightgray}elect2019: & \cellcolor{lightgray}John Kasich &  &  $3/4$ & \cellcolor{lightgray} $0/2$\\
        $\text{ChronoGPT-Instruct}_{2020}$  & \textcolor{blue}{\textbf{Bill Clinton}}  & \textcolor{blue}{\textbf{George W.}} & George W.  & \textcolor{blue}{\textbf{Donald Trump}}   & Bill Clinton & \cellcolor{lightgray}George W. &  &  $3/5$ & \cellcolor{lightgray} $0/1$\\
        $\text{ChronoGPT-Instruct}_{2021}$  & \textcolor{blue}{\textbf{Bill Clinton}}  & \textcolor{blue}{\textbf{George W.}} & George W.  & \textcolor{blue}{\textbf{Donald Trump}}   & \textcolor{blue}{\textbf{Joe Biden}}  &\cellcolor{lightgray} Joe Biden &  &  $4/5$ & \cellcolor{lightgray} $0/1$\\
        $\text{ChronoGPT-Instruct}_{2022}$  & \textcolor{blue}{\textbf{Bill Clinton}}  & \textcolor{blue}{\textbf{George W.}} & George W.  & \textcolor{blue}{\textbf{Donald Trump}}   & \textcolor{blue}{\textbf{Joe Biden}}  &\cellcolor{lightgray} Joe Biden &  &  $4/5$ & \cellcolor{lightgray} $0/1$\\
        $\text{ChronoGPT-Instruct}_{2023}$  & \textcolor{blue}{\textbf{Bill Clinton}}  & \textcolor{blue}{\textbf{George W.}} & George W.  & \textcolor{blue}{\textbf{Donald Trump}}   & \textcolor{blue}{\textbf{Joe Biden}}  &\cellcolor{lightgray} Joe Biden &  &  $4/5$ & \cellcolor{lightgray} $0/1$\\
        $\text{ChronoGPT-Instruct}_{2024}$  & \textcolor{blue}{\textbf{Bill Clinton}}  & \textcolor{blue}{\textbf{George W.}} & George W.  & \textcolor{blue}{\textbf{Donald Trump}}   & \textcolor{blue}{\textbf{Joe Biden}}  & Joe Biden &  &  $4/6$ & \cellcolor{lightgray} ---\\
        \hline
        \multicolumn{8}{l}{$\text{ChronoGPT-Instruct}_{1999}$ through $\text{ChronoGPT-Instruct}_{2024}$} & $67/83$ & \cellcolor{lightgray}$0/73$\\
        \bottomrule
    \end{tabular}
    }
    \caption{Next-Token Predictions of U.S. Presidents using ChronoGPT-Instruct}
    \label{tab:presidents_chronogpt}
        \bigskip
    \scriptsize
This table presents ChronoGPT-Instruct’s next-token predictions for prompts listing the incoming U.S. president alongside the three most recent predecessors. The model is tasked with predicting the name of the most recent president, which appears as the final missing entry in the sequence. The input prompt is structured as follows:

\begin{quotation}
``U.S. Presidents in chronological order:

Took office in \{year$_{p-3}+1$\}: President \{name$_{p-3}$\}

Took office in \{year$_{p-2}+1$\}: President \{name$_{p-2}$\}

Took office in \{year$_{p-1}+1$\}: President \{name$_{p-1}$\}

Took office in \{year$_{p}+1$\}: President \rule{2cm}{0.1mm}"
\end{quotation}

The table shows the predictions generated by each model in the ChronoGPT-Instruct series. Each prediction consists of exactly two tokens, selected deterministically by choosing the most probable token at each step. Gray shading indicates prompts referencing years beyond the model’s knowledge cutoff, including elections in which the president-elect had not yet assumed office. Correct predictions are highlighted in blue. For comparison, outputs from GPT-2, GPT-2 XL (released in 2019), Llama-3.2-3B-Instruct (released in 2023), and Qwen-1.5-1.8B-Chat (released in 2024) are also included.
\end{table}

\begin{table}[!htb]
    \resizebox{\textwidth}{!}{
    \begin{tabular}{lccccccccc}
        \multicolumn{10}{l}{Panel A: Input prompts} \\
        \toprule
        Event year & \multicolumn{6}{c}{Input prompt} & & \multicolumn{2}{c}{Correct output}\\
        \midrule
        2001 & \multicolumn{6}{l}{The Sarbanes-Oxley Act was introduced in response to the 2001 Enron \rule{2cm}{0.1mm}} & & \multicolumn{2}{c}{scandal}  \\
        2003 & \multicolumn{6}{l}{In 2003, a major public health crisis was the outbreak of the virus known as \rule{2cm}{0.1mm}}  & & \multicolumn{2}{c}{SARS} \\
        2008 & \multicolumn{6}{l}{In 2008, the global economy was dominated by the subprime mortgage \rule{2cm}{0.1mm}}  & & \multicolumn{2}{c}{crisis}  \\
        2016 & \multicolumn{6}{l}{In 2016, market volatility increased surrounding the general vote known as the Brexit \rule{2cm}{0.1mm}}  & & \multicolumn{2}{c}{referendum} \\
        2020 & \multicolumn{6}{l}{In 2020, the global economy was devastated by the health crisis known as the ``\rule{2cm}{0.1mm}}  & & \multicolumn{2}{c}{COVID/coronavirus} \\
        2022 & \multicolumn{6}{l}{In 2022, a major milestone for generative AI was marked by the release of the AI chatbot known as ``\rule{2cm}{0.1mm}}  & & \multicolumn{2}{c}{ChatGPT}  \\
        \bottomrule
        \\
        \multicolumn{10}{l}{Panel B: Output predictions} \\
        \toprule
        & \multicolumn{6}{c}{Event year} & & \multicolumn{2}{c}{Accuracy}\\
                \cmidrule{2-7} \cmidrule{9-10}
         & 2001 & 2003 & 2008 & 2016 & 2020 & 2022 & & Pre-cutoff & Post-cutoff \\
        \midrule
        Correct output & scandal & SARS & crisis & referendum & COVID/coronavirus & ChatGPT\\
        \midrule
        GPT-2              & \textcolor{blue}{\textbf{scandal}.\textbackslash n}  & chikung  & \textcolor{blue}{\textbf{crisis}, which}  & vote, with  &\cellcolor{lightgray}Great Recession."  &\cellcolor{lightgray} The AI Bot &  & $2/4$ & \cellcolor{lightgray}$0/2$\\
        GPT-2 XL              & \textcolor{blue}{\textbf{scandal}, in}  & \textcolor{blue}{\textbf{SARS},}   & \textcolor{blue}{\textbf{crisis}. The}  & . The UK &\cellcolor{lightgray}Great Recession."   &\cellcolor{lightgray} Alexa," &  & $3/4$ & \cellcolor{lightgray}$0/2$\\
        Llama-3.2-3B-Instruct   & \textcolor{blue}{\textbf{scandal}, which} & \textcolor{blue}{\textbf{SARS} (} & \textcolor{blue}{\textbf{crisis}, which} & \textcolor{blue}{\textbf{referendum}. The} & \textcolor{blue}{\textbf{COVID}-19} & LLaMA &  & $5/6$ & \cellcolor{lightgray}---\\  
        Qwen-1.5-1.8B-Chat   & \textcolor{blue}{\textbf{scandal}, which} & \textcolor{blue}{\textbf{SARS}.} & \textcolor{blue}{\textbf{crisis}. The} & \textcolor{blue}{\textbf{referendum}. The} & \textcolor{blue}{\textbf{COVID}-1} & \textcolor{blue}{\textbf{ChatGPT}} &  & $6/6$ & \cellcolor{lightgray}---\\    
        \midrule
        $\text{ChronoGPT-Instruct}_{\text{Realtime}}$ & \cellcolor{lightgray} ment Act, & \cellcolor{lightgray} the ``H & \cellcolor{lightgray} market, which & \cellcolor{lightgray} -Elli & \cellcolor{lightgray} H1N & \cellcolor{lightgray} AI Assistant" & & --- & \cellcolor{lightgray} 0/6 \\
        \midrule
        $\text{ChronoGPT-Instruct}_{1999}$   & \cellcolor{lightgray}ies Act, & \cellcolor{lightgray}the ``V & \cellcolor{lightgray}market, which & \cellcolor{lightgray}. The market & \cellcolor{lightgray} World Health Organization & \cellcolor{lightgray} Chatbot"  &  & --- & \cellcolor{lightgray}$0/6$\\
        $\text{ChronoGPT-Instruct}_{2000}$   & \cellcolor{lightgray}ment Act, & \cellcolor{lightgray}the Ebola virus & \cellcolor{lightgray}market, which & \cellcolor{lightgray}. The market & \cellcolor{lightgray} Asian Crisis."
        & \cellcolor{lightgray} Chatbot"  &  & --- & \cellcolor{lightgray}$0/6$\\
        $\text{ChronoGPT-Instruct}_{2001}$   & ment Crisis in & \cellcolor{lightgray}the ``Asian & \cellcolor{lightgray}market, which & \cellcolor{lightgray}. & \cellcolor{lightgray}Asian flu." & \cellcolor{lightgray} The Chatbot  &  & $0/1$ & \cellcolor{lightgray}$0/5$\\
        $\text{ChronoGPT-Instruct}_{2002}$   & \textcolor{blue}{\textbf{scandal}. It} & \cellcolor{lightgray}the ``H & \cellcolor{lightgray}market, which & \cellcolor{lightgray} . The market & \cellcolor{lightgray} Asian flu." & \cellcolor{lightgray} AI-1  &  & $1/1$ & \cellcolor{lightgray}$0/5$\\
        $\text{ChronoGPT-Instruct}_{2003}$   & \textcolor{blue}{\textbf{scandal}. It} & \textcolor{blue}{the \textbf{SARS}} & \cellcolor{lightgray}market, which & \cellcolor{lightgray} vote. & \cellcolor{lightgray} Asian flu." & \cellcolor{lightgray} Chatbot".  &  & $2/2$ & \cellcolor{lightgray}$0/4$\\
        $\text{ChronoGPT-Instruct}_{2004}$   & \textcolor{blue}{\textbf{scandal}. It} & \textcolor{blue}{\textbf{SARS},} & \cellcolor{lightgray}market, which & \cellcolor{lightgray} hip. & \cellcolor{lightgray} Asian flu." & \cellcolor{lightgray} Chatbot".  &  & $2/2$ & \cellcolor{lightgray}$0/4$\\
        $\text{ChronoGPT-Instruct}_{2005}$   & \textcolor{blue}{\textbf{scandal}. It} & \textcolor{blue}{\textbf{SARS},} & \cellcolor{lightgray}market, which & \cellcolor{lightgray} . & \cellcolor{lightgray} Asian flu." & \cellcolor{lightgray} Chatbot".  &  & $2/2$ & \cellcolor{lightgray}$0/4$\\
        $\text{ChronoGPT-Instruct}_{2006}$   & \textcolor{blue}{\textbf{scandal}. It} & \textcolor{blue}{\textbf{SARS}.} & \cellcolor{lightgray}market, which & \cellcolor{lightgray} . & \cellcolor{lightgray} Asian flu." & \cellcolor{lightgray} Chatbot".  &  & $2/2$ & \cellcolor{lightgray}$0/4$\\
        $\text{ChronoGPT-Instruct}_{2007}$   & \textcolor{blue}{\textbf{scandal}. It} & \textcolor{blue}{\textbf{SARS},} & \cellcolor{lightgray}market, which & \cellcolor{lightgray} . & \cellcolor{lightgray} Asian flu." & \cellcolor{lightgray} Chatbot".  &  & $2/2$ & \cellcolor{lightgray}$0/4$\\
        $\text{ChronoGPT-Instruct}_{2008}$   & \textcolor{blue}{\textbf{scandal}. It} & \textcolor{blue}{the \textbf{SARS}} & \textcolor{blue}{\textbf{crisis}, which} & \cellcolor{lightgray} . & \cellcolor{lightgray} Asian flu." & \cellcolor{lightgray} Chatbot".  &  & $3/3$ & \cellcolor{lightgray}$0/3$\\
        $\text{ChronoGPT-Instruct}_{2009}$   & \textcolor{blue}{\textbf{scandal}. It} & \textcolor{blue}{the \textbf{SARS}} & \textcolor{blue}{\textbf{crisis}, which} & \cellcolor{lightgray} . & \cellcolor{lightgray} SARS" & \cellcolor{lightgray} Chatbot".  &  & $3/3$ & \cellcolor{lightgray}$0/3$\\
        $\text{ChronoGPT-Instruct}_{2010}$   & \textcolor{blue}{\textbf{scandal}. It} & \textcolor{blue}{the \textbf{SARS}} & \textcolor{blue}{\textbf{crisis}, which} & \cellcolor{lightgray} vote. & \cellcolor{lightgray} Spanish flu" & \cellcolor{lightgray} The Language Model  &  & $3/3$ & \cellcolor{lightgray}$0/3$\\
        $\text{ChronoGPT-Instruct}_{2011}$   & \textcolor{blue}{\textbf{scandal}. It} & \textcolor{blue}{the \textbf{SARS}} & \textcolor{blue}{\textbf{crisis}, which} & \cellcolor{lightgray} vote. & \cellcolor{lightgray} Asian flu." & \cellcolor{lightgray} The Language Model  &  & $3/3$ & \cellcolor{lightgray}$0/3$\\
        $\text{ChronoGPT-Instruct}_{2012}$   & \textcolor{blue}{\textbf{scandal}, which} & \textcolor{blue}{the \textbf{SARS}} & \textcolor{blue}{\textbf{crisis}, which} & \cellcolor{lightgray} vote. & \cellcolor{lightgray} Asian flu" & \cellcolor{lightgray} The Language Model  &  & $3/3$ & \cellcolor{lightgray}$0/3$\\
        $\text{ChronoGPT-Instruct}_{2013}$   & \textcolor{blue}{\textbf{scandal}, which} & \textcolor{blue}{the \textbf{SARS}} & \textcolor{blue}{\textbf{crisis}, which} & \cellcolor{lightgray} vote. The & \cellcolor{lightgray} Asian flu" & \cellcolor{lightgray} The Turing Test  &  & $3/3$ & \cellcolor{lightgray}$0/3$\\
        $\text{ChronoGPT-Instruct}_{2014}$   & \textcolor{blue}{\textbf{scandal}, which} & \textcolor{blue}{the \textbf{SARS}} & \textcolor{blue}{\textbf{crisis}, which} & \cellcolor{lightgray} of the year & \cellcolor{lightgray} Spanish flu" & \cellcolor{lightgray} The Turing Test  &  & $3/3$ & \cellcolor{lightgray}$0/3$\\
        $\text{ChronoGPT-Instruct}_{2015}$   & \textcolor{blue}{\textbf{scandal}, which} & \textcolor{blue}{the \textbf{SARS}} & \textcolor{blue}{\textbf{crisis}, which} & \cellcolor{lightgray} -Elli & \cellcolor{lightgray} Spanish flu" & \cellcolor{lightgray} AI Language Model  &  & $3/3$ & \cellcolor{lightgray}$0/3$\\
        $\text{ChronoGPT-Instruct}_{2016}$   & \textcolor{blue}{\textbf{scandal}. It} & \textcolor{blue}{the \textbf{SARS}} & \textcolor{blue}{\textbf{crisis}, which} & . The European & \cellcolor{lightgray} Asian flu" & \cellcolor{lightgray} The Chatbot  &  & $3/4$ & \cellcolor{lightgray}$0/2$\\
        $\text{ChronoGPT-Instruct}_{2017}$   & \textcolor{blue}{\textbf{scandal}, which} & \textcolor{blue}{the \textbf{SARS}} & \textcolor{blue}{\textbf{crisis}, which} & \textcolor{blue}{\textbf{referendum}. The} & \cellcolor{lightgray} Asian flu" & \cellcolor{lightgray} The Chatbot  &  & $4/4$ & \cellcolor{lightgray}$0/2$\\
        $\text{ChronoGPT-Instruct}_{2018}$   & \textcolor{blue}{\textbf{scandal}. It} & \textcolor{blue}{the \textbf{SARS}} & \textcolor{blue}{\textbf{crisis}, which} & \textcolor{blue}{\textbf{referendum}. The} & \cellcolor{lightgray} Asian flu" & \cellcolor{lightgray} The Chatbot  &  & $4/4$ & \cellcolor{lightgray}$0/2$\\
        $\text{ChronoGPT-Instruct}_{2019}$   & \textcolor{blue}{\textbf{scandal}, which} & \textcolor{blue}{the \textbf{SARS}} & \textcolor{blue}{\textbf{crisis}, which} & \textcolor{blue}{\textbf{referendum}. The} & \cellcolor{lightgray} H1N & \cellcolor{lightgray} The Chatbot  &  & $4/4$ & \cellcolor{lightgray}$0/2$\\
        $\text{ChronoGPT-Instruct}_{2020}$   & \textcolor{blue}{\textbf{scandal}, which} & \textcolor{blue}{\textbf{SARS}-} & \textcolor{blue}{\textbf{crisis}, which} & \textcolor{blue}{\textbf{referendum}. The} & Spanish flu" & \cellcolor{lightgray} AI Language Model  &  & $4/5$ & \cellcolor{lightgray}$0/1$\\
        $\text{ChronoGPT-Instruct}_{2021}$   & \textcolor{blue}{\textbf{scandal}, which} & \textcolor{blue}{\textbf{SARS} in} & \textcolor{blue}{\textbf{crisis}, which} & \textcolor{blue}{\textbf{referendum}.} & \textcolor{blue}{\textbf{coronav}}  & \cellcolor{lightgray} AI Assistant"  &  & $5/5$ & \cellcolor{lightgray}$0/1$\\
        $\text{ChronoGPT-Instruct}_{2022}$   & Corporation collapse, & \textcolor{blue}{the \textbf{SARS}} & \textcolor{blue}{\textbf{crisis}, which} & \textcolor{blue}{\textbf{referendum}. The} & \textcolor{blue}{\textbf{coronav}}  & AI Assistant"  &  & $5/6$ & \cellcolor{lightgray}---\\
        $\text{ChronoGPT-Instruct}_{2023}$   & \textcolor{blue}{\textbf{scandal}, which}  & \textcolor{blue}{\textbf{SARS},} & \textcolor{blue}{\textbf{crisis}, which} & \textcolor{blue}{\textbf{referendum}. The} & \textcolor{blue}{\textbf{coronav}} & \textcolor{blue}{\textbf{ChatGPT}}  & & $6/6$ & \cellcolor{lightgray}---\\
        $\text{ChronoGPT-Instruct}_{2024}$   & \textcolor{blue}{\textbf{scandal}. It}  & \textcolor{blue}{\textbf{SARS},} & \textcolor{blue}{\textbf{crisis}, which} & \textcolor{blue}{\textbf{referendum}. The} & \textcolor{blue}{\textbf{COVID}-} & \textcolor{blue}{\textbf{ChatGPT}}  & & $6/6$ & \cellcolor{lightgray}---\\
        \hline
        \multicolumn{7}{l}{$\text{ChronoGPT-Instruct}_{1999}$ through $\text{ChronoGPT-Instruct}_{2024}$}  & & $76/80$ & \cellcolor{lightgray}$0/76$\\
        \bottomrule
    \end{tabular}
    }    \caption{Next-Token Predictions of Major Events using ChronoGPT-Instruct}
    \label{tab:majorevents_chronogpt}
        \bigskip
    \scriptsize
This table presents ChronoGPT-Instruct’s next-token predictions for prompts describing major historical events across a range of years. Panel A displays the input prompts for each event, and Panel B shows the corresponding predictions produced by each model in the ChronoGPT-Instruct series. Each prediction consists of exactly three tokens, selected deterministically by choosing the most probable token at each step. Gray shading indicates prompts that reference events beyond the model’s knowledge cutoff. Correct predictions are highlighted in blue. For comparison, outputs from GPT-2, GPT-2 XL (released in 2019), Llama-3.2-3B-Instruct (released in 2023), and Qwen-1.5-1.8B-Chat (released in 2024) are also included.
\end{table}

While we take great care in curating our training data to ensure it includes only information believed to be available as of a specific date, the process is not immune to error. Such errors may originate in either the pretraining or finetuning datasets. In the case of pretraining data, inaccuracies in recorded publication dates, such as those introduced when printed materials are digitized via optical character recognition (OCR) and assigned incorrect timestamps, can lead to the inadvertent inclusion of information that was not actually available at the intended time. For finetuning, classification was performed using GPT-4.1, which may misidentify whether certain prompt-response pairs reflect knowledge from before or after the year 2000. These issues can result in lookahead bias, unintentionally exposing the model to information that should not have been available at the specified training cutoff.

For the pretraining dataset, \cite{he2025chronologically} test for leakage in the ChronoGPT series using textual sequences involving U.S. presidents and major events from various years, and find no evidence of leakage. In this section, we replicate their validation exercise for ChronoGPT-Instruct to test whether there is additional leakage introduced in the instruction finetuning process.

Table \ref{tab:presidents_chronogpt} presents the test involving U.S. presidents and Table \ref{tab:majorevents_chronogpt} presents the test involving major events. In both tables, the gray-shaded area in the top-right indicates predictions in the post-knowledge cutoff period, and the non-shaded lower-right denotes predictions strictly in the pre-knowledge cutoff period. Correct predictions are highlighted in blue. 

Within the respective knowledge window of each version of ChronoGPT-Instruct, ChronoGPT-Instruct correctly makes a majority of predictions (67 out of 83) for the U.S. presidents test and correctly makes most predictions (76 out of 80) for the major events test. The high accuracy in the pre-knowledge cutoff period reflects the quality and temporal relevance of the ChronoGPT-Instruct's knowledge. In contrast, during the post-cutoff period represented by the gray-shaded area, none of the ChronoGPT-Instruct models correctly predict any future president or major event. Overall, these findings validate that the textual data used to train or finetune our chronologically consistent models contains no evidence of leakage.

\subsection{Prompt-Based Trading Portfolios}

While \cite{he2025chronologically} highlight the potential of ChronoGPT to generate profitable signals through embeddings following \citet{chen2023expected}, it remains an open question how well directly prompting a chronologically consistent model can serve a similar purpose. \citet{lopez-lira_can_2023} demonstrate that prompting LLMs can yield robust trading signals for a limited time period after the model's knowledge cutoff, without introducing lookahead bias. With ChronoGPT-Instruct, we extend this line of inquiry by conducting prompt-based portfolio construction over a substantially longer time horizon, from January 2007 to July 2023.

To operationalize this investigation, we apply the following prompt to financial news headlines at the stock-day level, for all stocks with news coverage, similar to \citet{lopez-lira_can_2023}:

{\ttfamily
\noindent\{
\begin{adjustwidth}{2em}{2em}
\#\#\# Instruction:\\
Classify this news headline as either FAVORABLE, or UNFAVORABLE, or UNCLEAR for the stock price of {company}. \\
\#\#\# Input:\\
\{headlines\} \\
\#\#\# Response:
\end{adjustwidth}
\noindent\}
}


We then form portfolios based on the LLM's response. The stock is assigned to the favorable news ($H$) or unfavorable news ($L$) portfolio based on the first word generated by the LLM. If the first word is neither ``favorable'' nor ``unfavorable'', the stock is assigned to the ``unclear'' portfolio. A long-short portfolio ($H-L$) is formed from longing the favorable news portfolio and shorting the unfavorable news portfolio.

We evaluate this strategy on $\text{ChronoGPT-Instruct}_{\text{Realtime}}$, which is trained and tuned entirely on data before the prediction year, and is free from lookahead bias.
For comparison, we also include results from Qwen-1.5-1.8B-Chat, Llama-3.2-3B-Instruct and Llama-3.2-1B-Instruct. The model closest in size is Qwen-1.5-1.8B-Chat, which has a 20\% larger parameter count. The next larger model, Llama-3.2-3B-Instruct, has twice the parameter count. Both models are trained on substantially more data than ChronoGPT-Instruct.

\begin{table}[!htb]
\small
\begin{center}
\bigskip
\begin{tabularx}{1.01\textwidth}{@{\hskip\tabcolsep\extracolsep\fill}l*{6}R}
\toprule
& \multicolumn{3}{c}{$\text{ChronoGPT-Instruct}_{\text{Realtime}}$}  & \multicolumn{3}{c}{Qwen-1.5-1.8B-Chat}
\\
\cmidrule(lr){2-4} \cmidrule(lr){5-7} 
& Mean & SD & SR & Mean & SD & SR  \\
\midrule
Unfavorable ($L$) & 1.35 & 24.53 & 0.05 & -1.67 & 25.90 & -0.06 \\
Unclear & 0.14 & 29.03 & 0.00 & 7.20 & 78.95 & 0.09 \\
Favorable ($H$) & 9.51 & 23.33 & 0.41 & 10.55 & 22.72 & 0.46 \\ 
\cmidrule(lr){1-1} \cmidrule(lr){2-4} \cmidrule(lr){5-7} 
$H-L$ & 8.17 & 8.63 & 0.95 & 12.21 & 8.00 & 1.53 \\
\midrule
& \multicolumn{3}{c}{Llama-3.2-3B-Instruct} & \multicolumn{3}{c}{Llama-3.2-1B-Instruct}\\
\cmidrule(lr){2-4} \cmidrule(lr){5-7}
& Mean & SD & SR & Mean & SD & SR \\
\midrule
Unfavorable ($L$)    & -1.11 & 25.68 & -0.04 & 4.67 & 23.64 & 0.20 \\
Unclear   & 6.71 & 23.17 & 0.29 & 14.01 & 23.04 & 0.61  \\
Favorable ($H$)    & 13.46 & 23.31 & 0.58 & 7.31 & 23.91 & 0.31 \\
\cmidrule(lr){1-1} \cmidrule(lr){2-4} \cmidrule(lr){5-7} 
$H-L$   & 14.58 & 8.31 & 1.76 & 2.64 & 6.91 & 0.38 \\
\bottomrule
\end{tabularx}
\end{center}
\caption{Performance of Prompt-Based Trading Portfolios}
\label{tab:LS}
\footnotesize
\bigskip
This table presents annualized performance metrics (mean return, standard deviation, and Sharpe ratio) for portfolios sorted by the LLM's direct response categorizing the news headlines as favorable news ($H$), unfavorable news ($L$), or unclear. The $H-L$ row represents a strategy of going long on the portfolio of stocks classified as having favorable news ($H$) and short on those with unfavorable news ($L$). 
All values are in percentage points except the Sharpe ratios. All portfolios are equal-weighted and rebalanced daily. Data spans January 2007–July 2023.
\end{table}

\begin{figure}[!htb]
    \begin{center}
    
    \includegraphics[width=1\linewidth]{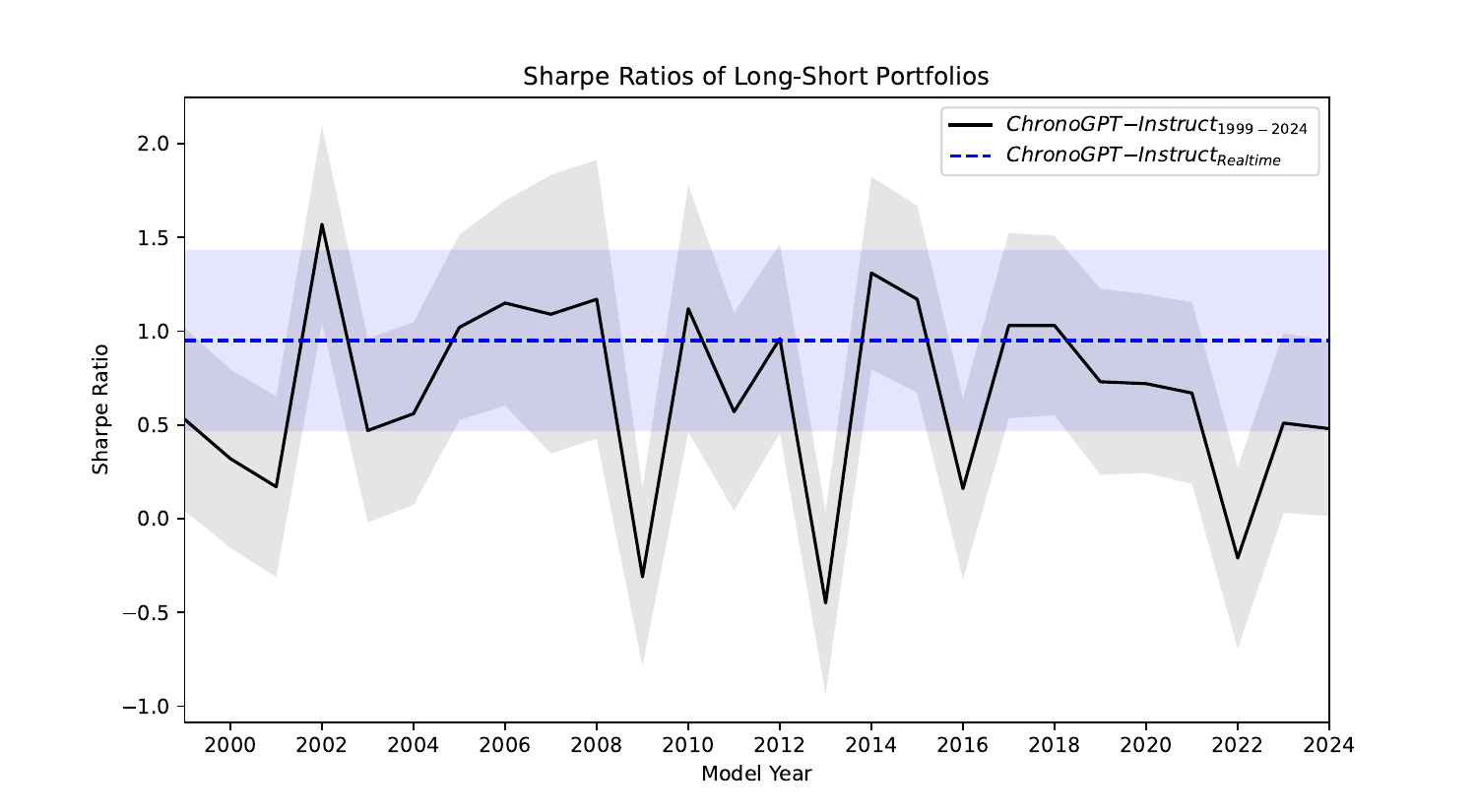}
    
    \end{center}
    \caption{Portfolio Performance across ChronoGPT-Instruct Vintages}
    \label{fig:SR_time}
    \bigskip
    \small
    This figure illustrates the Sharpe ratios of long-short portfolios constructed using predictions by ChronoGPT-Instruct, with each model pretrained on text data up to the time points indicated on the x-axis. The blue dashed line represents the performance of the ChronoGPT-Instruct$_\text{Realtime}$, using the model from the year before the prediction year. The shaded regions represent the 95\% confidence intervals.  
\end{figure}

Table \ref{tab:LS} presents the results of the prompt-based trading portfolios. The realtime $\text{ChronoGPT-Instruct}$ model achieves a Sharpe ratio of 0.95, outperforming Llama-3.2-1B-Instruct while underperforming the comparatively larger models, Qwen-1.5-1.8B-Chat and Llama-3.2-3B-Instruct.

If a chronologically consistent model matches an inconsistent counterpart in architecture and training, then matching return performance would imply that the inconsistent model’s predictability does not rely on leakage. In our setting, return predictability reflects two forces: language capability, which typically increases with parameter count and training tokens, and lookahead bias. ChronoGPT-Instruct is smaller and trained on fewer tokens than Qwen-1.5-1.8B-Chat and Llama-3.2-3B-Instruct, so its performance serves as a conservative lower bound for the leakage-free component. Comparing $\text{ChronoGPT-Instruct}_{\text{Realtime}}$'s Sharpe ratio of 0.95 to that of Qwen-1.5-1.8B-Chat's (1.53) and Llama-3.2-3B-Instruct's (1.76), at least 54\% to 62\% of the apparent return predictability persists in the absence of data leakage. The remaining gap in Sharpe ratios (e.g., between 0.95 and 1.76) likely reflects a combination of differences in model capacity and lookahead bias in the comparison model.

\cite{he2025chronologically} pose a critical question: while later models demonstrate more up-to-date knowledge and improved language understanding as they are trained on more data over time, does this translate into economic gains? To test this, they evaluate the trading performance of the entire series of chronologically consistent models. Our analysis using instruction-tuned versions of those same models reveals a distinct performance pattern in Figure \ref{fig:SR_time} that corroborates their ``envelope'' phenomenon.

The primary explanation for this envelope pattern is twofold. First, the results demonstrate that lookahead bias is modest. If significant lookahead bias were present, the final model, with the most comprehensive knowledge, would have been the top performer across all periods. The fact that it is not indicates that its 'knowledge of the future' does not grant it an unfair advantage for past data.

Second, and more critically for this task, further improvements in knowledge and generic language ability add only marginal value. The strong performance of even the earliest models shows that the performance frontier is reached relatively quickly. A key factor here is temporal alignment, that a model's calibration to the specific language, vocabulary, and market narratives of its own era. Expressions such as ``meme stocks'' or ``supply-chain disruptions'' carry period-specific meanings, and interpreting them through a future-biased lens can misalign signals and erode predictive accuracy.

However, we identify a notable distinction in our results: the performance gap between the real-time model and the other vintage models is less pronounced than in the original study. While the real-time model still outperforms the average vintage, its advantage is diminished.

We hypothesize that this attenuated effect stems from the IFT process applied across all vintages. Because the IFT dataset is temporally fixed to a pre-1999 period, all models may become less temporally aligned with their respective pretraining eras. By finetuning every vintage on linguistic patterns from a single, static historical era, the unique temporal signature learned during pretraining is likely diluted. This ``anchoring'' to a past linguistic style may weaken the specialized advantage of the real-time model, thereby compressing the performance difference across the different vintages.



\section{Conclusion}\label{sec:conclusion}

We address lookahead bias in LLM-based predictions by releasing the first chronologically consistent instruction-following language models whose training corpora are explicitly time-stamped. For example, $\text{ChronoGPT-Instruct}_{1999}$ is trained and subsequently instruction-finetuned exclusively on text available up to 1999, giving researchers over two decades of true out-of-sample period for evaluation.

While we acknowledge that temporal constraints necessarily limit model performance compared to contemporary alternatives, ChronoGPT-Instruct models offer a conservative lower bound for quantifying lookahead bias. A practical way to gauge the extent of lookahead bias is to compare ChronoGPT-Instruct with similarly sized but temporally inconsistent models such as Qwen-1.5-1.8B-Chat or Llama-3.2-3B-Instruct. Instead of achieving state-of-the-art performance, our aim is to provide an easy-to-use, replicable benchmark for quantifying lookahead bias in a wide range of prompt-based prediction tasks.

\clearpage
\newpage
\onehalfspacing
\bibliographystyle{jf}
\bibliography{chrono_instruct}

@techreport{ludwig2025large,
  title={Large language models: An applied econometric framework},
  author={Ludwig, Jens and Mullainathan, Sendhil and Rambachan, Ashesh},
  year={2025},
  institution={National Bureau of Economic Research}
}

@article{he2025chronologically,
  title={Chronologically Consistent Large Language Models},
  author={He, Songrun and Lv, Linying and Manela, Asaf and Wu, Jimmy},
  journal={arXiv preprint arXiv:2502.21206},
  year={2025}
}

@article{chen2023expected,
	title = {Expected Returns and Large Language Models},
	url = {https://papers.ssrn.com/abstract=4416687},
	language = {en},
	urldate = {2024-07-15},
	author = {Chen, Yifei and Kelly, Bryan T. and Xiu, Dacheng},
	journal = {SSRN Electronic Journal},
	year = {2023},
}

@article{lopez-lira_can_2023,
	title = {Can {ChatGPT} Forecast Stock Price Movements? Return Predictability and Large Language Models},
	shorttitle = {Can {ChatGPT} {Forecast} {Stock} {Price} {Movements}?},
	url = {http://arxiv.org/abs/2304.07619},
	doi = {10.48550/arXiv.2304.07619},
	urldate = {2024-07-14},
	author = {Lopez-Lira, Alejandro and Tang, Yuehua},
	month = sep,
	year = {2023},
        journal = {SSRN Electronic Journal},
}

@article{chang2023ai,
  title={AI democratization, return predictability, and trading inequality},
  author={Chang, Anne and Dong, Xi and Martin, Xiumin and Zhou, Changyun},
  journal={Available at SSRN 4543999},
  year={2023}
}

@article{chen2025chatgpt,
  title={ChatGPT and DeepSeek: Can they predict the stock market and macroeconomy?},
  author={Chen, Jian and Tang, Guohao and Zhou, Guofu and Zhu, Wu},
  journal={arXiv preprint arXiv:2502.10008},
  year={2025}
}

@article{glasserman2023assessing,
  title={Assessing look-ahead bias in stock return predictions generated by gpt sentiment analysis},
  author={Glasserman, Paul and Lin, Caden},
  journal={arXiv preprint arXiv:2309.17322},
  year={2023}
}

@article{sarkar2024lookahead,
  title={Lookahead bias in pretrained language models},
  author={Sarkar, Suproteem K and Vafa, Keyon},
  journal={Available at SSRN},
  year={2024}
}

@techreport{jha2024chatgpt,
  title={ChatGPT and corporate policies},
  author={Jha, Manish and Qian, Jialin and Weber, Michael and Yang, Baozhong},
  year={2024},
  institution={National Bureau of Economic Research}
}

@article{engelberg2025entity,
  title={Entity Neutering},
  author={Engelberg, Joseph and Manela, Asaf and Mullins, William and Vulicevic, Luka},
  journal={Available at SSRN},
  year={2025}
}

@article{sarkar_storieslm_2024,
	title = {{StoriesLM}: A Family of Language Models With Time-Indexed Training Data},
	author = {Sarkar, Suproteem},
	year = {2024},
        journal = {SSRN Electronic Journal},
}

@article{lv2025sell,
  title={Do Sell-side Analyst Reports Have Investment Value?},
  author={Lv, Linying},
  journal={arXiv preprint arXiv:2502.20489},
  year={2025}
}

@book{raschka2024build,
  title={Build a large language model (from scratch)},
  author={Raschka, Sebastian},
  year={2024},
  publisher={Simon and Schuster}
}

@article{wang2022self,
  title={Self-instruct: Aligning language models with self-generated instructions},
  author={Wang, Yizhong and Kordi, Yeganeh and Mishra, Swaroop and Liu, Alisa and Smith, Noah A and Khashabi, Daniel and Hajishirzi, Hannaneh},
  journal={arXiv preprint arXiv:2212.10560},
  year={2022}
}

@article{lambert2024tulu,
  title={Tulu 3: Pushing frontiers in open language model post-training},
  author={Lambert, Nathan and Morrison, Jacob and Pyatkin, Valentina and Huang, Shengyi and Ivison, Hamish and Brahman, Faeze and Miranda, Lester James V and Liu, Alisa and Dziri, Nouha and Lyu, Shane and others},
  journal={arXiv preprint arXiv:2411.15124},
  year={2024}
}

\end{document}